\documentclass{article}

\PassOptionsToPackage{numbers, compress}{natbib}

\usepackage[final]{nips_2018}

\usepackage[utf8]{inputenc} 
\usepackage[T1]{fontenc}    
\usepackage{hyperref}       
\usepackage{url}            
\usepackage{booktabs}       
\usepackage{amsfonts}       
\usepackage{nicefrac}       
\usepackage{microtype}      
\usepackage{fullpage,enumitem,amsmath,amssymb,graphicx}
\usepackage{float}
\title{Skeleton-based Coherence Modeling in Narratives}

\author{
  Nishit Asnani \\
  Department of Computer Science\\
  Stanford University\\
  Stanford, CA \\
  \texttt{nasnani@stanford.edu} \\
  \And
  Rohan Badlani \\
  Department of Computer Science\\
  Stanford University\\
  Stanford, CA \\
  \texttt{rbadlani@stanford.edu} \\
}

\begin{document}

\maketitle

\begin{abstract}
  Modeling coherence in text has been a task that has excited NLP researchers since a long time. It has applications in detecting incoherent structures and helping the author fix them. There has been recent work in using neural networks to extract a skeleton from one sentence, and then use that skeleton to generate the next sentence for coherent narrative story generation. In this project, we aim to study if the consistency of skeletons across subsequent sentences is a good metric to characterize the coherence of a given body of text. We propose a new Sentence/Skeleton Similarity Network (SSN) for modeling coherence across pairs of sentences, and show that this network performs much better than baseline similarity techniques like cosine similarity and Euclidean distance. Although skeletons appear to be promising candidates for modeling coherence, our results show that sentence-level models outperform those on skeletons for evaluating textual coherence, thus indicating that the current state-of-the-art coherence modeling techniques are going in the right direction by dealing with sentences rather than their sub-parts.  
  
\end{abstract}

\section{Introduction}
With the success of neural networks in a myriad variety of natural language processing (NLP) tasks over the last few years, the community is naturally drifting towards more nuanced understanding of natural language, and tackling the problems that emerge therein. If we consider natural language understanding, the trend seems to be from syntactical understanding (parsing, named entity recognition) to a deeper semantic understanding (granular sentiment analysis, intent prediction). Modeling coherence of a given piece of text lies in the latter class of problems, and is quite challenging because even the average person is not perfect at qualitatively identifying coherence quality. 

Broadly stated, coherence in a text is what leaves a well-tied impression of the themes discussed and the arguments made in the text in the mind of the reader. An incoherent piece of text is harder to tie together for the reader, might be inconsistent or have irrelevant chunks of information, or in the worst case, have no logical progression of ideas. According to Wikipedia, coherence is achieved by a combination of the use of linguistic elements in a logical tense structure, as well as "presuppositions and implications connected to general world knowledge." 

Over the last two decades, various attempts have been made by NLP researchers to model coherence in texts, many of them working in English texts.\cite{li2016neural, doi:10.1162/coli.2008.34.1.1} These attempts use varying word representations, varying models to classify a sentence as coherent or incoherent with respect to its context, and define their task in a couple of different ways. Coherence is a fundamental part of a well-written narrative, and thus there have been recent efforts to generate narratives based on some measure of preserving coherence. In particular, we build upon the idea of 'skeletons' presented by Jingjing Xu et. al\cite{Skeleton-Based-Generation-Model} at EMNLP 2018, which uses preservation of sentence skeletons for generating coherent narratives. We then hypothesize that if their model is performing well in generation, it should also be a good model for detecting incoherent sentences in a given piece of text - thus, we turn it from a generative model to a discriminative one. Finally, we build upon this idea and propose a Sentence/Skeleton Similarity Network to evaluate the similarity between two consecutive sentences in a given text using either skeletons or raw sentences. We conduct a series of structured experiments that help us analyze the suitability of modeling coherence in terms of skeletons and compare its performance with coherence modeled using raw sentences. 

\section{Related work}
Modeling coherence in any given text segment is essential for summarization, question-answering and analysis of the text. There have been many approaches to model coherence in a given text segment. Barzilay and Lapata \cite{doi:10.1162/coli.2008.34.1.1} presented a novel framework for representing and measuring local coherence which uses entity-grid representation of discourse, which captures patterns of entity distribution in a text. Arguably this has been the most successful approach to automatically learning text coherence is the entity grid, which relies on modelling patterns of distribution of entities across multiple sentences of a text. Jurafsky et al \cite{li2016neural} presented domain independent neural models of discourse coherence that are capable of measuring multiple aspects of coherence in existing sentences. 

Cervone et. al \cite{DBLP:journals/corr/abs-1806-08044} propose to augment the original grid document representation for dialogues with the intentional structure of the conversation. They show that their models outperform the original grid representation on both text discrimination and insertion, the two main standard tasks for coherence assessment across three different dialogue datasets, confirming that intents play a key role in modelling dialogue coherence. Motivated by this intuition, we think that key phrases within the a sentence and not just the intent play a major role in determining coherence within the text. 

Jingjing Xu et. al \cite{Skeleton-Based-Generation-Model} provide an input-skeleton-output pipeline, wherein at test time they extract the key concepts (entities, relations, events etc.) from the input sentence as a skeleton, and use that skeleton to generate the output sentence. This uses the intuition that composing an essay or narrating a story as a human, we work with key ideas and expand those into coherent text. Thus, the authors rely on the premise that coherent extension of a narrative is dependent on having consistent skeletons across consecutive sentences. The skeleton extraction module is pre-trained using a sentence compression task. The input-to-skeleton and skeleton-to-output modules is trained along with finetuning the skeleton extraction module in a reinforcement learning setting. The input-to-skeleton and skeleton-to-output components are both LSTM based Seq2Seq models. 

However, Jingjing Xu et. al \cite{Skeleton-Based-Generation-Model} focus on the problem of generating narrative stories that are coherent from an input consisting of a short description of the event or scene. Most of the state of art techniques like proposed by Jain et al. \cite{DBLP:journals/corr/JainAMSLS17}, Fan et al. \cite{DBLP:journals/corr/abs-1805-04833} for sentence generation use Sequence2Sequence models that we studied in CS224N. The problem with formulating a narrative generation of story with models like seq2seq is that it is hard to model the semantic dependency among sentences. 

The concept of skeletons introduced by Jingjing Xu et. al is fairly new and to the best of our knowledge, there has not been any work around utilizing skeletons to model coherence of text. The authors  have demonstrated that the proposed new skeleton-based model for generating coherent narrative stories performs quite well as compared to baselines. Thus, motivated by our intuition that key phrases within sentences should help identify coherence, we turn the problem addressed by Jingjing Xu et. al from a generative model to a discriminative one. In this paper, we extend their work further by accessing the quality of the text using skeletons as a measure for coherence and try to extend this work for suggesting rephrasing of a incoherent sentence or text to enhance coherence. 

\begin{center}
\begin{figure}
\centering
\includegraphics{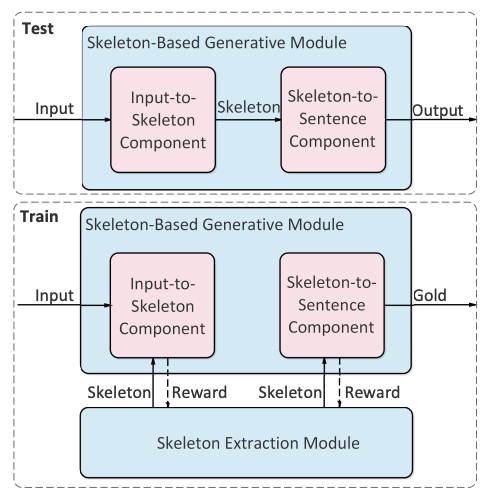}
\caption{Skeleton-based Model for generating narratives, with test phase and training phase clearly demarcated.}
\end{figure}
\end{center}

\section{Approach}
In this project, we set out to explore the following hypothesis: since using the first sentence's skeleton to generate the second sentence leads to coherent story generation, we evaluate whether skeleton similarity across sentences is a good measure for coherence modeling. In this direction, we implement a skeleton similarity network (SSN), which takes in the skeletons from two sentences, constructs dense vector representations for them, and evaluates how similar those representations are. We compare this model to the following:

\begin{itemize}
    \item A baseline where the SSN is replaced with non-parametric measures of similarity, namely cosine similarity and (negative) Euclidean distance. 
    \item The original model with the exception that the SSN is trained on sentences instead of skeletons, in order to be able to compare the utility of skeletons with respect to sentences in coherence modeling. 
\end{itemize}

\begin{figure}
\centering
\includegraphics[width=0.9\textwidth]{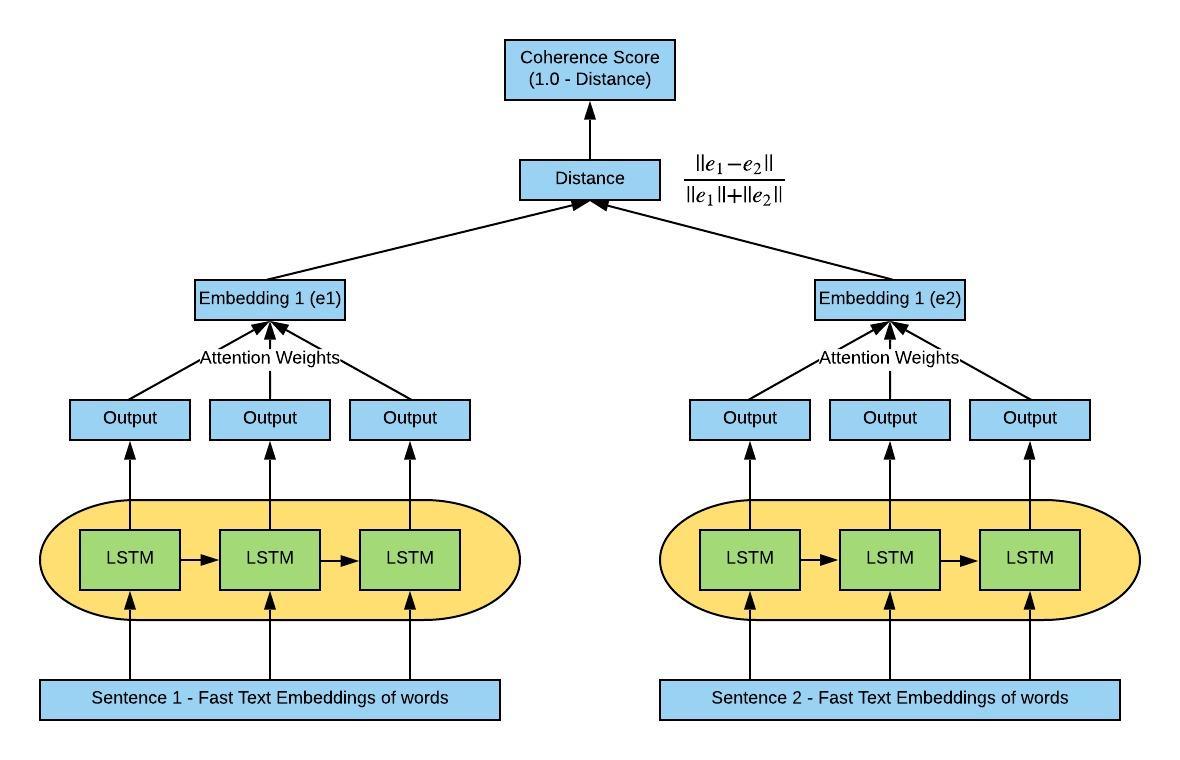}
\caption{Skeleton Similarity Network. For the experiments where the attention layer is removed, we use the output from the final encoder LSTM as the sentence embedding.}
\end{figure}

For each of the two skeletons (or sentences), our model takes their word embeddings in sequence and passes them through a \textbf{Long Short Term Memory (LSTM)} Network to obtain sentence embeddings. Finally, the normalized L2 distance between the two sentence embeddings is used as a distance measure (1 - distance is used as similarity measure). The model is penalized on \textbf{contrastive loss} on the distance prediction. We implement this Siamese Network based Skeleton/Sentence Similarity Network by extending some of the implementation provided by Neculoiu et al.\cite{neculoiu2016learning}. 

The loss calculation is using the following set of equations. Given 2 sentences, $s_{1}$ and $s_{2}$, let their embeddings after the LSTM and potentially the self-attention layer be $e_{1}$ and $e_{2}$. The energy ($E_w$) is defined as normalized cosine similarity between the 2 embeddings:

\begin{equation}
    E_w = \frac{cosine(e_{1}, e_{2})}{ \mid {e_{1}}\mid \mid e_{2} \mid}
\end{equation}

The contrastive loss for the similar and dissimilar
cases are given by:

\begin{equation}
    L_{pos}(e_{1}, e_{2}) = \frac{1}{4} * (1- E_w)^2
\end{equation}

\begin{equation}
    L_{neg}(e_{1}, e_{2}) = \begin{cases}
                    E_w^2 , \text{if} E_w < m\\
                        0 , else \end{cases}
\end{equation}

\begin{equation}
    L_w^i (e_{1}^i, e_{2}^i, y^i) = y^i L_{pos}(e_{1}^i, e_{2}^i) + (1 - y^i) L_{neg}(e_{1}^i, e_{2}^i)
\end{equation}

\begin{equation}
    Loss = \sum_{i=1}^{N} L_w^i (e_{1}^i, e_{2}^i, y^i)
\end{equation}

We augment this basic architecture by introducing \textbf{FastText} word embeddings\cite{bojanowski2016enriching} at the input, instead of making the model learn fresh word embeddings for this task. This is useful since a sentence skeleton is generally not a contiguous chunk from the original sentence, and thus most word embedding methods don't make sense to be trained from such data (since word order and relationships become unclear). This also enables the model to use very high quality word embeddings to train the rest of the architecture better. 

In addition, we introduce \textbf{self attention} on the LSTM sentence / skeleton encoder outputs in one of our models, to further improve the quality of sentence-level embeddings. We implement this ourselves, since off-the-shelf attention wrappers on LSTM cells on Tensorflow were not working as expected. This works as follows: 

\begin{align}
    \begin{split}
        c_{i} & = \sum{W_{ij}o_{j}} \\
        \alpha & = softmax(c) \\
        o' & = \sum{\alpha_{i}o_{i}} 
    \end{split}    
\end{align}

Here, $c$ is the context vector derived from the outputs $o$, $\alpha_{i}$ are the normalized attention weights, and $o'$ is the final output, which is a convex combination of the outputs, weighted by the normalized attention weights. 

This mechanism is inspired by Luong et.al. \cite{luong2015effective}, but since we only have an encoder here and no decoder, the attention weights are derived from the encoder outputs and are used on the encoder outputs as well.  

\section{Experiments}
\subsection{Data and necessary preprocessing}
We use the story telling dataset by \cite{huang2016visual} for the purposes of evaluation of the our proposed technique. The dataset includes 40153, 4990, and 5054 stories for training, validation, and testing, respectively each consisting of maximum 6 sentences. We use the skeleton-based model described \cite{Skeleton-Based-Generation-Model} to train the model and compute the skeletons from the trained model for each of the training, validation and test sets. 

The Sentence/Skeleton Similarity Network (SSN) as proposed above relies on data that can tell the distinction between a pair of sentences. Since we already know the ground truth on the data, we prepare the training, testing and validation datasets for the SSN model using the ground truth. We prepare 2 types of datasets for testing the SSN - one that is sentence level and the other which is paragraph level. The construction of the dataset is descibed below:

\begin{enumerate}
    \item \textbf{Sentence Pairs}: We consider 2 consecutive sentences within the actual story and treat them as being similar (label 1). For the first sentence within this we pick a random sentence from any other story and treat this sentence pair as non-similar (label=0).
    
    \item \textbf{Story Pairs}: We consider consecutive sentences within the actual story as sentences that are in the same order and call this is as ordered story. Now we jumble the sentences within a story to form the jumbled story. 
\end{enumerate}

\subsection{Evaluation Metrics}
We evaluate our models on three distinct metrics:
\begin{itemize}
    \item \textbf{Incoherent sentence pair detection}: Given a pair of consecutive sentences and a pair of randomly sampled sentences (with one sentence common across both the pairs), is the model able to distinguish between the two (by assigning a higher score to the former)?
    \item \textbf{Incoherent story pair detection}: Given the original story as a list of sentences and a copy of the same story with sentence order randomized, is the model able to distinguish between the two (by assigning a higher coherence score to the former)?
    \item \textbf{Classifying sentence pairs}: Given two sentences, is the model able to detect if they are consecutive (by implicitly measuring coherence) or not?
\end{itemize}
The first two are metrics that are commonly used in coherence modeling literature. Since our model spits out a sentence similarity score as its output, given a pair of sentences, we average out these similarities across all consecutive sentence pairs in a story to arrive at a story coherence score. This score is then used for comparison using the second metric described above. For the third metric, we evaluate the model by labeling every coherence score prediction over 0.5 as 1 and under 0.5 as zero, and then compare this with the ground truth. 

\subsection{Evaluation Details}
The model hyper-parameters for both skeleton-based model as well as the SSN are available below

\begin{itemize}
    \item \textbf{Skeleton Extraction Module}: Since the \cite{Skeleton-Based-Generation-Model} have done comprehensive analysis on the hyper-parameter tuning for this model, we re-use the same hyper parameters as suggested by them. Following table summarizes:
    
    \begin{center}
        \begin{tabular}{|c c|}
        \hline
         Hyperparameter & Value  \\
         \hline
         Hidden state dim of RNN & 128\\
         \hline
         Embedding Dim of wordvecs & 50\\
         \hline
         Mini-batch Size & 10\\
         \hline
         Learning Rate & 0.6\\
         \hline
         Init accumulator value for Adagrad & 0.1\\
         \hline
         Random uniform initialization RNN & 0.02\\
         \hline
         Gradient clipping & 2.0\\
         \hline
    \end{tabular}
    
    \end{center}
    
    \item \textbf{FastText Word Embeddings}: Rather than using pre-trained word vectors, we train word vectors for our data using the open source implementation of FastText provided by facebook research. \cite{bojanowski2016enriching}
    
    \begin{center}
        \begin{tabular}{|c c|}
        \hline
         Hyperparameter & Value  \\
         \hline
         Embedding Dim of wordvecs & 100\\
         \hline
    \end{tabular}
    
    \end{center}
    
    \item \textbf{Skeleton/Sentence Similarity Network}: We train our SSN model separately for sentences and skeletons, but we use the same hyper-parameters for each since we want to compare their performance. 
    
    \begin{center}
        \begin{tabular}{|l c|}
        \hline
         Hyperparameter & Value  \\
         \hline
         Hidden Units & 50 \\
         \hline
         Dropout Keep Prob & 1.0\\
         \hline
         L2 regularizaion lambda & 0.0\\
         \hline
         Mini-batch Size & 64\\
         \hline
         Epochs & 100\\
         \hline
         Stacked LSTM Layers & 2-3\\
         \hline
    \end{tabular}
    \end{center}
    
\end{itemize}

\section{Results and Analysis}
First, we show our results on using sentence/skeleton similarity with non-parametric similarity measures like cosine and Euclidean distance, which are stacked on top of sentence/skeleton embeddings obtained by averaging BERT word embeddings for each word in the sentence/skeleton. As we had shown in the milestone, cosine similarity works better than Euclidean distance on the sentence order metric (metric 1 described above): 

\begin{center}
 \begin{tabular}{|l c|} 
 \hline
 Technique & Sentence order accuracy (in \%)  \\ [0.5ex] 
 \hline
 BERT embedding means for sentence + Euclidean distance  & 68.3  \\ 
 \hline
 BERT embedding means for skeleton + Euclidean distance  & 59.3  \\
 \hline
 BERT embedding means for sentence + cosine similarity  & 71.9  \\
 \hline
 BERT embedding means for skeleton + cosine similarity  & 61.6  \\
 \hline
 
\end{tabular}
\end{center}

Next, we compare neural approaches to sentence similarity with each other on all the three accuracy calculation metrics (in \%) - identifying ordered (and therefore coherent) sentence pairs, identifying ordered / coherent stories, and classifying sentence pairs as being coherent or incoherent.  

\begin{center}
 \begin{tabular}{|l c c c|} 
 \hline
 Technique & Sentence order & Story order & Pair classification \\ [0.5ex] 
 \hline
 SSN-3 on sentences  & 92.9  & 69.6 & 82.2 \\
 \hline
 SSNA-2 on sentences & 92.3  & 68.0 & 81.4 \\
 \hline 
 SSN-3 on skeletons  & 84.2  & 62.9 & 73.8 \\
 \hline
 SSNA-2 on skeletons & 84.5  & 62.3 & 74.5 \\
 \hline
\end{tabular}
\end{center}

\textbf{Firstly}, the above results show that neural approaches to modeling coherence work better than non-parametric approaches, even after using BERT embeddings, by a huge margin. Thus, it is worthwhile to pursue coherence modeling by training fresh neural networks despite the prominent recent advances in contextual word representations. 

As we can see from the results above, the SSN models give a reasonably good performance in the task of detecting if 2 given sentences (or respectively skeletons) are similar to each other or not. This can be seen from the Pair Classification metric in the table above. Both the sentence and skeleton based model reach a performance of 82\% and 74\% respectively. Even with these SSN models, we observe that the empirical performance of sentences in identifying the quality of the text outperforms that of skeletons.

The \textbf{second} important finding is that even with SNN models that show good performance in detecting similarity between 2 sentences or skeletons, empirical results show that the sentence based techniques perform better than skeletons in the task of coherence detection between 2 sentences. This can be seen from both Sentence order and Story order metrics (metric definitions in Sec 4.2). As can be seen, the sentence-based models show a performance of around 92\% whereas the skeleton-based models only show a performance of around 84\% for Sentence-order metric. We see a similar trend in the Story-order metric as well where sentences outperform skeletons. 

Although, we had expected that since skeletons are key words within the document, they should be better candidates for coherence within a document. However, empirically we observe that skeletons are in fact not a good choice for coherence modelling. There can be attributed to 2 reasons:

\begin{itemize}
    \item Since the skeletons are themselves detected from a highly complex neural model (input-to-skeleton component in Fig 1), the quality of the skeletons will significantly affect their performance in terms of coherence modelling.
    
    \item Moreover, since the similarity network tries to feed off on contextual occurrence of words within the input sentence or skeletons, it is natural that the sentences, which have complete set of words and in a order, would have a better performance in detecting whether 2 sentences are similar or not, however in the case of skeletons, the relatively very short lengths without any order leads to reduction in the performance of the SSN.
\end{itemize}

While this bodes positively for existing approaches on coherence modeling (since almost all of them use sentence inputs), it is contrary to what we expected going into the project.

\textbf{Thirdly}, on closely analyzing the performance of the models across the sentence-order and story-order metrics, we find that both sentence-based and skeleton-based SSN models perform much better at sentence-level coherence detection than paragraph-level coherence detection. This can be attributed to the fact that the story telling dataset contains stories with a maximum sentence length of 6. So, when we compute a jumbled sequence of sentences from a story, it is quite likely that 2 consecutive sentences within the story end up together within the jumbled set and hence the way we evaluate the metric (which is finding if ordered story similarity > unordered story similarity), due to lack of absolute random orders, the models perform better at sentence metric than story metric. After doing some literature survery around this, we find that Jurafsky et al. \cite{li2016neural} suggest that datasets that contain paragraphs/stories with at least 16 sentences should be considered for story-order coherence evaluation. Our hypothesis and extension to this work is that when the models are evaluated on larger datasets like NTSB airplane accident reports and AP earthquake reports (as used by \cite{barzilay2008modeling}) or essay dataset\cite{ASAP_dataset} consisting of 500 words or more, our SSN models would perform similar on story-order metric and sentence-order metric, if not better.

\textbf{Lastly}, when we compare the performance of self-attention based models with models without attention, we observe that our implementation of self-attention, does not boost the performance significantly. Although, we had expected a bump in the performance of our SSN models, it turns out that self-attention yields similar performance on all 3 metrics especially in the case of sentences. It was our expectation that self-attention should be able to pick and focus on the right set of words within a sentence by placing more weight and hence more contribution towards the embedding but we do not observe this empirically. Although surprising at first, but one possible reason for this can be the fact that due to limited compute availability and time constraints, we decide to use a 2 layered stacked LSTM + self-attention SSN model whereas our SSN model without attention uses 3-layered stacked LSTM. However, since the 2-layered stacked LSTM + self-attention is able to perform as well as normal SSN model, our hope is that if we train a 3 layered stacked LSTM + self-attention SSN model with proper regularization, it will perform better than our normal 3-layered SSN. 

\section{Conclusion}
Modelling coherence is a challenging task since even the average human is not perfect at identifying coherence quality of a given text sequence. There have been quite a few approaches to modelling coherence and some of them include entity-grid based representation of discourse, domain independent neural models and recent success of utilizing intentional structure in coherence assessment. Motivated by the progress in this direction, through this paper we test if skeletons or key-phrases within a text is a good way to measure the coherence of the text. Building upon the ideas on skeleton detection for coherence story generation, we propose a novel architecture called Sentence/Skeleton Similarity Network (SSN) to identify the similarity between pair of sentences or skeletons. We show that our proposed neural model is much better than non-parametric similarity techniques like cosine and (negative) euclidean distance. 

After validating the performance of our proposed SSN, we set out to model coherence of a given text by the similarity between consecutive sentences within a text and compare that with similarity between consecutive skeletons. We use 2 different metrics for evaluation - one at sentence level and other at story level. Contradictory to our initial hypothesis and expectation, the empirical performance shows that sentence-based models consistently perform better across both metrics than the skeleton based models. After rigorous analysis, we conclude that the performance of skeleton detection strongly affects the performance of our SSN and the relative sizes and chaotic order of the words within skeletons leads to difficulty in modelling coherence using skeletons. We also show that self-attention mechanism in the SSN is able to place emphasis on the right words when applied to sentence and provide good overall performance when used on sentences. Thus, our final conclusion is that sentences are better for coherence modelling than skeletons and models like self-attention based SSN proposed in this paper provide good performance in detecting coherence quality of a given text.

\section{Future Work and Extensions}
We think the work in the paper can be further extentended and enhaced in the following ways:
\begin{itemize}
    \item Evaluation on longer length datasets like \cite{ASAP_dataset} and \cite{joty2018coherence} will provide a more conclusive result in terms of coherence detection on larger sequences of text. 
    
    \item We applied a very simplistic self-attention mechanism in this paper which can be extended to complicated ideas like the ones presented in Transformers. \cite{DBLP:journals/corr/VaswaniSPUJGKP17}
    
    \item Evaluating the proposed SSN in detecting incoherent sentences within a text document with 1-2 sentences incoherent rather than fully ordered or jumbled stories, essays or paragraphs. This can have significant applications in wide variety of domains.
 \end{itemize}

\section{Acknowledgements}
We would like to thank the entire Stanford University CS224N (Natural Language Processing with Deep Learning) Teaching Staff. Particularly, we really appreciate the guidance of Anand Dhoot, our mentor for the project who is immensely knowledgeable in the domain of NLP and guided us throughout the project. He helped us by providing potential reasons when our models did not perform well and suggested that we use FastText embeddings which are faster to compute that really helped us complete our experiments on time. We also appreciate Prof Manning and Abigail See (Head TA) for the class for providing us valuable learnings in the domain of NLP and helping us refine our project idea.

\section{Appendix}
\subsection{Code}
Our entire codebase has been made public and is available at \url{https://github.com/rohanbadlani/skeleton-based-coherence-modelling}. The \textit{master} branch consists of model without self-attention and the one with self-attention is in the \textit{self-attention} branch.

\pagebreak

\bibliography{references}
\bibliographystyle{unsrtnat}

\end{document}